# Flow Segmentation in Dense Crowds

Javairia Nazir
Fatima Jinnah Women University
Rawalpindi, Pakistan
javairianazir@gmail.com

Mehreen Sirshar
Fatima Jinnah Women University
Rawalpindi, Pakistan
msirshar@gmail.com

**Abstract:**

**Now a day's security is an important aspect to be considered, so a framework is proposed in this paper that is used to segment flow of dense crowds. The flow field that is generated by the movement in the crowd is treated just like an aperiodic dynamic system. On this flow field a grid of particles is put over for particle advection by the use of a numerical integration scheme. Then flow maps are generated which associates the initial position of the particles with final position. The gradient of the flow maps gives the amount of divergence of the neighboring particles. For forward integration and analysis forward FTLE is calculated and backward FTLE is also calculated it gives the Lagrangian coherent structures of the flow in crowd. LCS basically divides the flow in crowd into regions and these regions have different dynamics. These regions are then used to get the boundary in the different flow segments by using water shed algorithm. The experiment is conducted on the crowd dataset of UCF.**

## 1. Introduction:

In large crowd gatherings at events like religious festivals, concerts and football matches public safety is a big challenge. The movement of these crowds is in a confined space. Only the public safety organizations are not able solve these problems alone. In February 2004, 251 pilgrims had breathed their last at the Jamarat Bridge. Due to this, bridge was redesigned but on January 2006, almost 345 people again lost their lives at this bridge because of stampede.

During past years different algorithms were proposed and designed but the major problem about the systems lies in their instability to handle large crowds. When the density of crowd increases (as shown if Fig.1) the performance of these systems started degrading. This research is based on building a system which models the heavily dense crowded scenes. In the paper a term modelling is used which defines the segmentation of the dominant crowd flow.

In this research we are treating the moving crowd as aperiodic dynamic system. The dynamics reflect to the motion due to spatial temporal interactions between the individuals and with their physical environment. The motion patterns based on dynamics of motion are called as "flow segments". A framework for flow segmentation is proposed which uses the turbulence theory [10], and fluid dynamics [11]. In this approach we used coherent structures that basically exist in the fluid flow and they are determined by using particle advection of fluids**.** The flow is divided into different regions by Coherent structures and coherent behavior is shown by all particles which belong to the same region. This idea can also be related with the flow field that is generated by crowd. When CS is studied by using the quantities that are resulting from trajectories then they are called Lagrangian Coherent Structure (LCS).

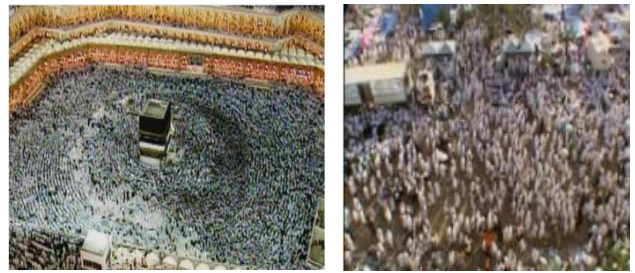

**Fig. 1**The example scenarios involving a large number of people

Different research has been performed to detect and track the individuals in crowds. A.Nijranjil Kumar [8] proposed a method of crowd behavior analysis by crowd analysis algorithm by first counting the people and then people tracking and crowd analysis. Ramin Mehran [5] uses videos in which particle grid was positioned on the image and moving particles are treated as individuals. Then force flow of the pixels is calculated in every frame. Vijay Mahandevan [10] identified the anomalies on spatio-temporal basis. The anomalies of temporal basis are equated for the events having low probability, and spatial anomalies deals with locations that are spatially abnormal which are determined by having saliency above some threshold. In Oluwatoyin P. Popoola [5] the behavior descriptor proposed in the study uses features that encode both local and global signatures of crowd interactions. In this research experiments are conducted on two crowd behavior datasets and the results shows that it is effective as compared to other methods and has good detection accuracy rates.

## 2. Methodology:

This section describes the details of crowd flow segmentation.
The first step is to consider the flow field between the consecutive frames. The flow fields which are calculated between the frames is then averaged to get one mean of the fields for the Mecca sequence. The next task is computing the particle advection. This is computed using the Lagrangian trajectory corresponding to the particles at the location of grid (a0,b0) and then by solving the differential equations $\frac{da}{dt}$= u (a,b,t) and $\frac{db}{dt}$ = v(a,b,t). Now for timet+T we have to calculate the trajectory.

Two flow maps are generated for the grid of these particles. In the first map a track is maintained showing the change in the x coordinate of particles y map keeps track that how the coordinate y particles are changing. At first these particles have their initial position on the map. Then these particles are advected using particle advection so the position of the particles also changes in the maps until end of the integration time length T.

Finite time exponents are determined by the initial position and by using length of integration of trajectories. The finite time Lyapunov exponentsare analyzed over the particle grid thus it produces a finite Lyapunov Exponent field. This field is computed using the flow map and it is a scalar field and it captures the dynamics of flow and also the geometry. The segmentation mechanism adopted comprises of two steps. The first step segment the crowd on the basis of FTLE field. In the other step those segments which have similar behavior in the boundary particles are merged. For segmentation we adopted water-shed algorithm, this is the algorithm that returns label matrix which identifies the watershed region.

The watershed algorithm uses the ridgeline of the image where light pixels represent the high elevations and the darks pixels are representing the low elevations. We first remove all the small segments less than 150 pixels this limit can also be changed. Then we remove all the segments where flow magnitude appears to be zero. These segments are called vacuum segment. The elements are labelled such that the elements which are labelled as zero do not belong to a unique watershed. Elements which are labelled as 1belongs to the first region of watershed and 2 for second region and so on. In the sequence we used, thousands of pilgrims around Kabba are circling anticlockwise. In this sequence the flow segment includes group of people as a whole, that are circlingand they belongs to the same segment, as all of them are indulged in accomplishment the same task. The framework used is shown in the Fig. 2

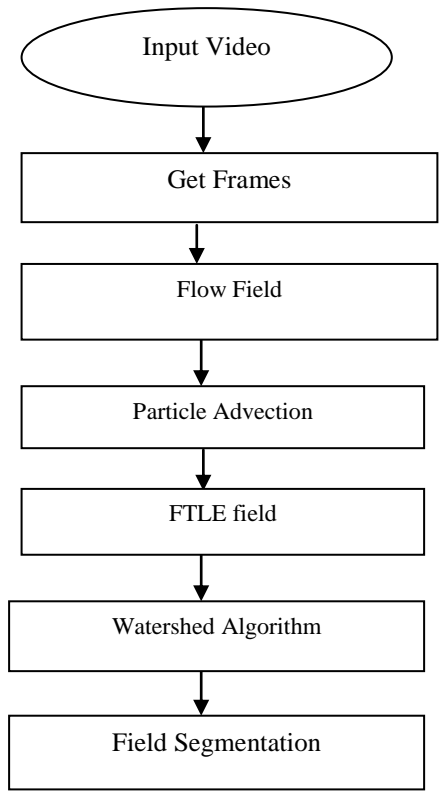

**Fig. 2** The flow of the proposed approach.

The sequence shown in Fig. 3 is taken as input as it involves a large number of people. Then Flow field is estimated. Then by particle advection flow maps are generated. At first we compute the forward integration by computing the forward FTLE in which we also remove the boundary data of that may be due to the noisy optical flow field at the image boundary. Then smooth the computed FTLE by a Guassian. For reverse integration and analysis the same steps are repeated. Then water shed algorithm is applied to which makes the segmentation. The segmented image of the frames is output of the process.

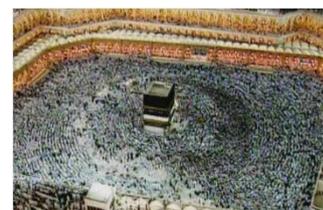

Fig. 3 The scenario involving
a number of people circling around Kabba

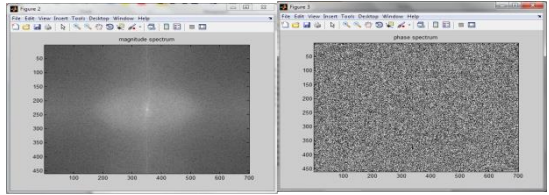

**Fig.**4 The magnitude and phase spectrum of flow calculated by FFT

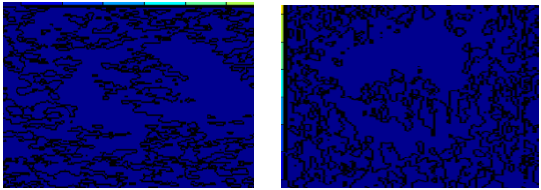

**Fig.5** Flow Maps (a) X-Flow map (b) Y-Flow Map

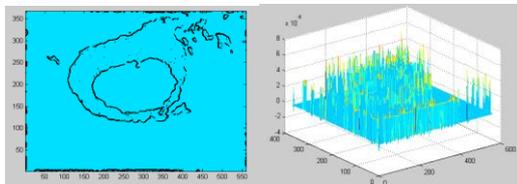

**Fig. 6 (a)** Spatial gradient of x-flow map (b) plot of peaks in FTLE

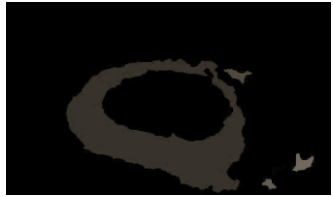

**Fig. 7** Segmented Image

### 3. Testing/ Results:

Center of research in computer vision has different dataset that are used in the crowd analysis. In this paper UCF crowd dataset is used for the segmentation of densely crowded scenes. The approach is tested on the videos that represent the high density of crowds. The flow field is computed for each video by using the methods described above. After computing the forward and backward particle advection the watershed algorithm is applied which makes the segments of the flow field. For the sequence we used as shown in Fig. 1 there is a dense crowd. This sequence shows a large group of people moving anticlockwise Kabba. The entire group of people that are circling in the anticlockwise direction belongs to the same flow segment as all of them are executing the same task.

The resulting image we obtain is a segmented image which contain flow segment of the crowd. This algorithm is applicable only to sequences having a dense crowd. This is applicable for the crowd analysis having a large number of people or a dense crowd. The crowd dataset available by UCF can be used for the crowd flow segmentation. This is important part in the behavior analysis of the crowded scenes and can be used for the instability or abnormal behavior in the crowd. This is not applicable for the anomaly detection in less number of people because it is designed according to sequences that involve dense crowd.

**Table 1. Comparison with other approaches**

| No. | Papers | Accuracy | Performance |
|---|---|---|---|
| 1. | Habib Ullah, 2012 | 90.9% | |
| 2. | Nan Li, 2011 | - | 0.987 |
| 3. | Lan Dong , 2007 | 74% | - |
| 4. | Oluwatoyin P. Popoola, 2012 | - | 0.935 |
| 5. | Ramin Mehran, 2009 | - | 0.786 |
| 6. | Axel Carlier , 2000 | - | 0.89 |
| 7. | Loris Nanni, 2013 | 76.02% | - |
| 8. | A.Niranjil Kumar, 2014 | 97.35% | - |
| 9. | Simon and Steven M. Seitz | - | 0.69 |
| 10 | Javairia Nazir | 70% | 0.5 |

## 4. Conclusion:

In this paper a framework is proposed that uses the Lagrangian particle dynamics to make the flow segmentation in crowds. Segmenting a dense crowd is useful in the analysis of the crowd behavior. This framework is use to segment the flow fields in a crowd by applying the particles advection, computing FTLE and watershed algorithm which segments the flow filed of the crowd. As in our sequence the people moving around Kabba are segmented as a common flow segment as they all are doing the similar activity. This is very important component for the analysis of instability in crowd. When there is instability in the crowd it can be detected by expanding this algorithm. When there is synthetic instability there is a change in the behavior of crowd flow. So by checking correspondence in segments with the learned set of segments instability is detected. Thus in future this algorithm can be used to analyze the abnormal behavior from normal in crowd scenes.